\documentstyle[eaclap]{article}

\title{Multilingual Sentence Categorization \\ according to Language
\thanks{This Paper is published in the {\it Proceedings of the
European Chapter of the Association for Computational Linguistics
SIGDAT Workshop ``From text to tags : Issues in Multilingual Language
Analysis''} held March 95 in Dublin.}}

\author{Emmanuel Giguet\\ GREYC --- CNRS URA 1526 --- Université de
Caen \\
  Esplanade de la Paix \\ 14032 Caen cedex --- France \\
   {\small e-mail\string: Emmanuel.Giguet@info.unicaen.fr}}

\begin{document}
\maketitle

\begin{abstract}

  Issues in sentence categorization according to language is
  fundamental for NLP, especially in document processing. In fact,
  with the growing amount of multilingual text corpus data becoming
  available, sentence categorization, leading to multilingual text
  structure, opens a wide range of applications in multilingual text
  analysis such as information retrieval or preprocessing of
  multilingual syntactic parser.

  The major difficulties in sentence categorization are convergence
  and textual errors. Convergence since dealing with short entries
  involve discarding languages from few clues. Textual errors since
  documents coming from different electronic ways may contain spelling
  and grammatical errors as well as character recognition errors
  generated by OCR.

  We describe here an approach to sentence categorization which has
  the originality to be based on natural properties of languages with
  no training set dependency. The implementation is fast, small,
  robust and textual errors tolerant. Tested for french, english,
  spanish and german discrimination, the system gives very interesting
  results, achieving in one test 99.4\% correct assignments on
  real sentences.

  The resolution power is based on grammatical words (not the most
  common words) and alphabet. Having the grammatical words and the
  alphabet of each language at its disposal, the system computes for
  each of them its likelihood to be selected. The name of the language
  having the optimum likelihood will tag the sentence --- but non
  resolved ambiguities will be maintained.  We will discuss the
  reasons which lead us to use these linguistic facts and present
  several directions to improve the system's classification
  performance.

  Categorization sentences with linguistic properties shows that
  difficult problems have sometimes simple solutions.

\end{abstract}

\section{Categorization according to Language}

\subsection{From Text Categorization \ldots}

Emergence of text categorization according to language came with the
need of processing texts coming from all over the world. The goal of
text categorization is to tag texts with the name of the language in
which they are written. Information retrieval is the main application
field.

To do this job, the traditionnal way is to exploit the difference
between letter combinations in different languages \cite{Cavnar94}.
For each language, the system computes from a training set a profile
based on frequency (or probability) of letter sequences. Then, for a
given text, it computes a profile and select the language which has
the closer profile.

While some text categorization systems give very good results, the
major problem is that their quality is entirely based on the training
set. Profiles require a lot of data to converge and building a large
representative training set is a real problem. Moreover, this method
assume that texts are monolingual and results will be affected when
dealing with multilingual texts. It does not care about natural
language properties : it only considers texts as streams of characters.
There is no linguistic justification.

\subsection{\ldots to Multilingual Sentence Categorization}
\label{subsec:tosentence}

Today, the problem is quiet different. Texts are more and more
multilingual (especially due to citations) and we don't have enough
tools to process them efficiently. Tagging sentences with the name of
their language solves this problem by switching each application in
function of the language. This affects the whole NLP, Information
retrieval is not the only field to be concerned: syntactic analysis and
every applications based on it are concerned, making study about one
particular language in multilingual texts without parasitic noise is
also possible.

Using the previous method is not possible because the sentence is a
too small unit to converge. The analysis method must be more
precise to reveal each possible change of language.

We remark that a change of language in a text could appear at each
change of sentence (more often paragraph) or in each included segment
via quotes, parenthesis, dashes or colons. We will call sentence the
traditionnal sentence but also each segment included in it.

\section{Multilingual Sentence Categorization}

Studying quantities of texts, we try to understand as well as possible
ways to discriminate languages. We present in this section the
results of our research which has been implemented and in the next
section, other directions which seems obviously promising.

\subsection{Grammatical Words as Discriminant}

In this section, we are going to motivate the reasons which lead us
to choose grammatical words as discriminant.

Grammatical words are proper to each language and are in a whole
different from one language to another. Moreover, they are short, not
numerous and we can easily build an exhaustive list. So, these words
can be use as discriminant of language. But can we use them as
discriminant of sentences?

Grammatical words in sentences represent on average about
50\% of words. They can't be omitted because they structure sentences
and make them understandable. Furthermore, relying on grammatical
words allows textual errors tolerance and foreign words import from
other languages (usual in scientific texts). It's also important to
note that foreign words import concerns nouns, verbs, adjectives but
never grammatical words.

These rules will allow us to categorize sentences which have enough
grammatical words but in short sentences (less than 10 words), there
are few grammatical words, and by the way, few clues. We must
introduce new knowledges to improve short sentences categorization.

\subsection{Using the Alphabet}

To improve categorization of short sentences, a simple way is the use
of the alphabet. Alphabets are proper to each language and even if
they have a great common part, some signs such as accents allows
discrimination between them. This is not the only way to improve
categorization and we will see in section \S\ref{sec:improving} other
possible issues.

\subsection{Notes}

\begin{itemize}
\item It is interesting that, using these knowledges, this system will
  be coherent with multilingual syntactic parsers which only rely on
  grammatical words and endings. So, the categorization system can
  constitute a switch for these parsers \cite{Vergne93,Vergne94}.

\item We can also remark that using grammatical words is different
  from using most common words. In fact, most common words require
  training set dependency and it is well known that a representative
  training set is very difficult to get. The number of words to hold
  is quiet subjective. Moreover, frequency is relative to texts, not
  to sentences.

\end{itemize}

\section{Improving Categorization}
\label{sec:improving}

There are two levels to improve sentences categorization: a level
below using words morphology and a level above using text structure.
These improvements haven't been implemented yet and will be the object
of further works.

\subsection{Knowledge upon Words Morphology}

Mainly two ways can be explore to improve categorization, using
natural languages properties:
\begin{itemize}
  \item Syllabation: the idea is to check the good syllabation of
    words in a language. It requires to distinguish first, middles and
    last syllabs. (Using only endings seems to be a possible way)

  \item Sequences of voyells or consonants: the idea is that these
    sequences are proper to each language.
\end{itemize}

\subsection{Using Text Structure}

When dealing with texts, we can also use heuristical knowledge about
text structure:

\begin{itemize}
  \item In a same paragraph, contiguous sentences are written in the
    same language

  \item Titles of a paragraph are written in the same language as
    their body

  \item Included blocks in a sentence (via parenthesis, \ldots) are
    written in the same language as the sentence.
\end{itemize}

An interesting tool to build is a general document structure
recognizer. Theoritical issues in this field are in progress
\cite{Lucas92,Lucas93} but as far as we know no implementation has
been done yet.

\section{Implementation}

The implementation of this research can be divided in two parts:
sentence tokenization and language classification.

\subsection{Sentence tokenization}

Sentence tokenization is a problem in itsef because documents may come
through different electronic ways. Also a sentence doesn't always start
with a capitalized letter and finish with a full stop (especially in
emails). Texts are not formated and miscellaneous characters can be
found everywhere.

Acronyms, abbreviations, full names and numbers increase the problem
by inserting points and/or spaces everywhere without following any
rule. But, no rule can ever exist in free style texts.

We wrote a robust sentence parser which solves the majority of these
cases, allowing us to categorize in good conditions multilingual
sentences.

\subsection{Language classification}

The realization simply implements the previous ideas.

To manage the possible points of change of language via included
segments (see section \S\ref{subsec:tosentence}),
the language classification procedure uses a recursive algorithm
to easily handle changes of context.

The classification principle is the following:

\begin{itemize}

\item For each word of the sentence:

\begin{itemize}
\item Checked whether the word belongs to the grammatical words list
  of some languages.
\item If so, incremented their likelihood to be selected.
\item Checked whether the word morphology lets think it belongs to some
  languages.
\item If so, incremented their likelihood to be selected.

\end{itemize}

\item Tag the sentence with the names of the languages which have the
  same and highest likelihood.

\end{itemize}

This algorithm has a linear complexity in time.

\section{Evaluation}

\subsection{The Test-Bed}

The test-bed set has been prepared to process French, English, Spanish
and German. We use dictionnaries to get the grammatical words of each
language (see table \ref{tab:number}) and their alphabet.

We decided to use different kinds of documents to test robustness,
speed, precision and textual errors tolerance.  So, we collected
scientific texts, emails and novels (see table \ref{tab:size}).

\begin{table}
  \begin{center}
\begin{tabular}{||l|c||}
\hline
Language & Grammatical Words \\
\hline
French   & 301 \\ \hline
English  & 186 \\ \hline
Spanish  & 204 \\ \hline
German   & 158 \\
\hline
\end{tabular}
  \end{center}
  \caption{Number of Grammatical Words}
  \label{tab:number}
\end{table}

\begin{table}
  \begin{center}

\begin{tabular}{||l|c||}
\hline
Language   & Number of    \\
of Corpus  & Sentences     \\ \hline
French     & 4502   \\ \hline
English    & 6735   \\ \hline
Spanish    & 94     \\ \hline
German     & 393    \\
\hline
\end{tabular}
  \end{center}
  \caption{Size of Corpus}
  \label{tab:size}
\end{table}

\subsection{Results}

The results we obtained were expected. They express the fact that
a sentence is usually written with grammatical words and that
grammatical words are totally discriminant for sentences of more than
8 words.

{}From 1 to 3 words, there are mainly total undeterminations.  In fact,
the corpus shows that we are processing included segments (via quotes
and parenthesis) and there are no grammatical words and few clues to
rely on. Deductions really start between 4 and 6 words. Here, sentences and
grammatical words appear but in few quantities to allow a perfect
deduction.

These results show that alphabets are not good enough to discriminate
short sentences. Methods described in \S\ref{sec:improving} must be
implemented to improve results in this case.

\begin{table}[h]
  \begin{center}
\begin{tabular}{||l|c|c|c||}
\hline
Language   & Min &  Decisive & Max \\
of Corpus  & Length &  Word  & Length \\ \hline
French     & 1 & 8 & 125\\ \hline
English    & 1 & 7 & 76 \\ \hline
Spanish    & 1 & 4 & 42 \\ \hline
German     & 1 & 5 & 66 \\
\hline
\end{tabular}
  \end{center}
  \caption{Isolation of a single language}
  \label{tab:words}
\end{table}

 In table \ref{tab:words}, with the french corpus, the program always
succeeds in isolating a single language for all the sentences containing
from 8 to 125 words.  For less than 8 words there are still
ambiguities or total undetermination.

\subsection{Errors}

Isolating a single language does not mean exactly isolating the right
language. The error rate is about 0.01\% and concerns very short
sentences ("e mail" where "e" is analysed as Spanish), a change of
language without quotes in a sentence or an unexpected language (the
Latin "Orbi et Urbi").

\section{Conclusion}

This classification method is based on texts observation and
understanding of their natural properties. It does not depend on
training sets and converges fast enough to achieve very
good results on sentences.

This tool is now a switch of Jacques Vergne's multilingual
syntactic parser (for french, english and spanish).

The aim of this paper is also to point that the more the linguistic
properties of the object are used, the best the results are.

\end{document}